# Vision-based Navigation and Obstacle Avoidance via Deep Reinforcement Learning

**Paul Blum**[1†]     **Peter Crowley**[1,2†]     **George Lykotrafitis**[1]

[1] University of Connecticut, Department of Mechanical Engineering, Storrs, CT, USA
[2] Boston University, Department of Mechanical Engineering, Boston, MA, USA

## Abstract

Development of navigation algorithms is essential for the successful deployment of robots in rapidly changing hazardous environments for which prior knowledge of configuration is often limited or unavailable. Use of traditional path-planning algorithms, which are based on localization and require detailed obstacle maps with goal locations, is not possible. In this regard, vision-based algorithms hold great promise, as visual information can be readily acquired by a robot's onboard sensors and provides a much richer source of information from which deep neural networks can extract complex patterns. Deep reinforcement learning has been used to achieve vision-based robot navigation. However, the efficacy of these algorithms in environments with dynamic obstacles and high variation in the configuration space has not been thoroughly investigated. In this paper, we employ a deep Dyna-Q learning algorithm for room evacuation and obstacle avoidance in partially observable environments based on low-resolution raw image data from an onboard camera. We explore the performance of a robotic agent in environments containing no obstacles, convex obstacles, and concave obstacles, both static and dynamic. Obstacles and the exit are initialized in random positions at the start of each episode of reinforcement learning. Overall, we show that our algorithm and training approach can generalize learning for collision-free evacuation of environments with complex obstacle configurations. It is evident that the agent can navigate to a goal location while avoiding multiple static and dynamic obstacles, and can escape from a concave obstacle while searching for and navigating to the exit.

---

[†] Authors made equal contribution on this work.

# 1      Introduction

The capability of robots to perform autonomous target-driven navigation in unknown dynamic environments is essential to integrating them into our daily lives. However, traditional path-planning algorithms are largely focused on generating optimal paths within static environments with known geometric configuration. One such popular set of algorithms is potential planners, which combine attractive and repulsive functions to draw the agent towards a goal while repelling it from obstacles (Khatib, 1990; Ge and Cui, 2000). A path-planning algorithm is considered probabilistically complete if as runtime goes to infinity, the probability of the planner finding a path from a start location to a target location converges to 1, assuming a feasible path exists between the specified start and target. Potential-based planners are simple to implement but are not probabilistically complete. Often, they struggle to navigate through narrow passages, and can become trapped at local minima inside of concave obstacles. Graph-based search algorithms such as A* (Hart et al., 1968) and RRT* (Karaman and Frazzoli, 2011) are probabilistically complete but must discretize the configuration space, ultimately limiting a robotic system. In addition, these methods assume a static environment and, like potential planners, require detailed obstacle maps. Online sensor-based planners such as Simultaneous Localization and Mapping (SLAM) have been developed to eliminate reliance on foreknown environment configurations (Durrant-Whyte and Bailey, 2006). SLAM uses LiDAR data to generate and update the obstacle map during navigation, allowing for adaptive control in dynamic environments. Like traditional methods, SLAM relies on precise localization and is computationally expensive in high-dimensional configuration spaces.

Learning methods, such as Deep Reinforcement Learning (DRL), offer a promising alternative approach to navigation of unknown and partially observable dynamic environments. DRL is a machine learning technique in which an agent learns through trial and error a behavioral policy, which maximizes the numerical reward emitted from its environment. The agent "learns" through a pre-defined number of training episodes in which it interacts with its environment to gain experience for improving its policy. DRL algorithms have had great success in mastering Markov decision processes in Atari games and the game of Go (Mnih et al., 2015, 2016; Silver et al., 2016, 2017).

Recently, the autonomous navigation problem has been modeled as a Markov decision process (MDP) to be solved using DRL. Deep Dyna-Q learning was used to model emergency room evacuation of an agent represented by a point in 2D space (Zhang et al., 2021). Dyna-Q is a DRL method in which model-free Q-Learning and model-based reinforcement learning are combined to learn the optimal action value function (Sutton and Barto, 2018). The authors trained a Deep Q-Network (DQN) to predict the value of moving in one of eight different directions based on the agent's position within the evacuation environment. The algorithm achieved time-efficient start-to-goal navigation from any initial position in obstacle-containing environments with favorable and aversion zones. This learning scenario is well-suited for a robotic navigation task within a static environment where the robot's absolute position is always known. However, a static environment, precise localization, and full knowledge of obstacle configurations are often unrealistic assumptions for a robotic system.

The asynchronous Deep Deterministic Policy Gradient (DDPG) algorithm was used to achieve mapless start-to-goal navigation (Tai et al., 2017). DDPG uses actor-critic reinforcement learning where the policy (actor) and value function (critic) are each defined by a Deep Neural Network (DNN) (Lillicrap et al., 2019). The actor DNN was trained to adjust a robot's velocity based on its state which is defined by sparse 180° laser range data, the current velocity, and the relative distance to the goal. While the DDPG-learned policy greatly outperformed a baseline SLAM motion planner, the approach still relies on information about the goal location. This assumption cannot be made for all target-driven navigation problems.

Use of camera images, rather than positional information, to inform navigation allows an agent to adapt to dynamic environments and learn behaviors based on visual elements and patterns. Indeed, deep supervised learning based on monocular camera images achieved collision-free path following of forest trails by an unmanned aerial vehicle (Giusti et al., 2016). Similarly, reinforcement learning was used to train a remote-controlled car for high-speed obstacle avoidance based on estimated depth from monocular images (Michels et al., 2005). The combination of deep visual perception with reinforcement learning can generate robust control policies for complex environments (Mnih et al., 2015). In one such DRL algorithm, researchers used the output of monocular RGB data fed through depth-predictive convolutional neural networks as input to a dueling double deep Q-network (D3QN) for end-to-end obstacle avoidance (Xie et al., 2017). The D3QN approach accelerated the learning rate over that of traditional deep Q-learning and performed well in unknown environments. The robot's objective was limited to obstacle avoidance along a pre-defined loop. Efficient start-to-goal navigation in complex environments requires that an optimal action be determined at each state. In contrast, the dueling network architecture is designed to generalize learning for MDPs in which computing the advantage function is unnecessary for many states (Wang et al., 2016). While others have used this method to achieve start-to-goal navigation in a simulated environment, an agent's ability to navigate to a target from any location was not proven, as the starting position was held constant (Ruan et al., 2019).

Similarly, DRL was used to train a Long Short-Term Memory (LSTM) computational graph for action-correlated collision prediction based on raw RGB camera image data (Kahn et al., 2018). During training, the LSTM weights were adjusted to minimize the mean squared error between predicted collision probabilities and labeled training data from previous experiences. The agent navigated by adjusting its steering angle to minimize expected collision probability. The algorithm outperformed double Q-learning, navigating without collision for 750 meters by the end of training compared to 250 meters with double Q-learning.

To extend vision-based DRL to autonomous cars, a photo-realistic driving-simulation engine called VISTA (Virtual Image Synthesis and Transformation for Autonomy) was developed for lane-following training (Amini et al., 2020). The VISTA-trained algorithm was able to drive a car along a previously unseen road while maintaining lane-stability. However, as in (Xie et al., 2017) and (Kahn et al., 2018), the agent's objective was not extended to dynamic obstacle avoidance or start-to-goal navigation.

In this work, we extend the deep Dyna-Q method to dynamic obstacle avoidance and target-driven navigation by using raw RGB images from a low-resolution onboard monocular camera as inputs to a robotic agent's control policy. Our navigation problem can be formalized as a partially observable MDP because the agent can never exactly determine to what state its environment will transition at a given time. Unlike previous DRL applications, our methods successfully achieve simultaneous start-to-goal navigation and complex obstacle avoidance based on visual information. We demonstrate the capabilities of our algorithm by training in several static and dynamic environments. Obstacles within these environments are respawned in random locations at the start of each training episode. We define a static environment as having obstacles that remain stationary throughout an episode, and a dynamic environment as one in which obstacles are moving during each episode. We note that this is in contrast to many vision-based DRL navigation algorithms, which define a dynamic environment as one in which obstacles can change positions between episodes but are not in motion as the robot navigates (Zhu et al., 2017; Kulhánek et al., 2019). The ability of our algorithm to learn robust navigation policies in time-variant (i.e. dynamic) environments sets it apart from much previous work.

In our first static environment, we train an agent to navigate to the exit in a room devoid of obstacles to validate our method. Next, we train an agent in two scenarios with a single static obstacle: one in which the obstacle is cylindrical, and the other in which the obstacle is a concave semi-circular wall. Navigating around and out of concave obstacles has proven difficult for classical path-planning algorithms like potential planners. This makes concave obstacle avoidance an important benchmark for our system to have achieved. Our final static training environment consists of three cylindrical obstacles. In the case of dynamic environments, we first train an agent to evacuate a room with a single randomly moving cylindrical obstacle. Finally, we train the agent to reach the exit of a room with three randomly moving cylindrical obstacles. The initial positions of all static obstacles, the movement of dynamic obstacles, and the locations of the exit are all randomly generated.

## 2   Methods

The purpose of this work is to identify an approach to obtain reliable navigation policies for camera-faring robots. We present a deep Dyna-Q learning algorithm that combines model-free and model-based reinforcement learning, $\varepsilon$-greedy exploration-exploitation, and the train-target DNN update method to train a robot to escape from a room based on camera image data. Specifically, a DQN is trained to approximate the optimal action value function for time-efficient evacuation based on monocular camera images. This section will describe the evacuation environment, formulate the reinforcement learning problem, and outline the Dyna-Q training process. In addition, we will briefly explain the structures of the DNNs used in this work, as well as the software used to simulate the training process.

We frame our procedure in a building-evacuation scenario, where an agent must learn to escape an enclosure by moving itself to the exit. Aside from its wide applicability, we study the building-evacuation scenario because it establishes a sufficiently complex navigation problem in a contained environment that incentivizes efficiency in planned trajectories.

### 2.1   Deep Reinforcement Learning Framework

Reinforcement learning algorithms are mathematical formulations where the objective is to accumulate maximal reward by selecting an optimal series of actions through an environment. In order to utilize such methods to train robotic navigation we must impose a reward function that appropriately models the task. Pursuant to our building-evacuation scenario, the agent should be incentivized to take actions that guide swift traversal to the exit. In this spirit, we devise a simple scheme that uses negative reward to penalize time spent in the environment. The ensuing function outputs zero reward when the agent reaches the exit and -0.1 reward for each time step.

In a reinforcement learning scenario, taking an action allows an agent to transition from one state in an environment to another. As such, the implicit usefulness of taking any action depends on the agent's current state. The notion that state-action transition pairs can be evaluated based on a perceived quality is fundamental in reinforcement learning. The quality of state-action pairs is quantified by the action-value function, commonly called the Q-value function. Formally, the value of taking action *a* from state *s* under policy $\pi$ is approximated by the Q-value $Q_\pi(s,a)$, which equals the expectation of cumulative episodic reward to be received after transitioning out of *s* via *a* described by the expression $Q_\pi(s,a) = \sum_{s',R} p(s',R \mid s,a)[R + \gamma v_\pi(s')]$ (Sutton and Barto, 2018). In this expression, p(*s'*, *R*|*s*, *a*) is the probability that the agent arrives in state *s'* and receives reward *R* after taking action *a* from state *s*, $\gamma \in [0,1]$ is the discount factor, and $v_\pi(s')$ is the expected cumulative reward, under policy $\pi$, from state *s'* over a finite horizon. The discount factor ($\gamma$) determines the degree to which future rewards are prioritized.

To maximize cumulative episodic reward, we seek to converge to a policy $\pi$ that suitably approximates the optimal policy (∗). This policy's learned action-value function would demonstrate proficient understanding of reward allocation in an environment, which could be utilized to make optimal decisions. This idea is embodied by the well-known Bellman Optimality Equation (Sutton and Barto, 2018):

$$Q_*(s,a) = \sum_{s',R} p(s',R \mid s,a) \left[ R + \gamma \max_{a'} Q_*(s',a') \right] \geq Q_\pi(s,a), \ \forall \pi. \tag{1}$$

In this equation, $Q_*(s,a)$ is the optimal action-value function and $a'$ is the action of greatest value from state $s'$. Q-learning (Watkins and Dayan, 1992) is a common way to find solutions to the Bellman Optimality Equation. It is an iterative method based on Q-value updates:

$$Q(s,a) = Q(s,a) + \alpha \left[ R + \gamma \max_{a'} Q(s',a') - Q(s,a) \right]. \tag{2}$$

In this equation, $\alpha$ is a configurable learning rate that limits how much the action-value function can change between updates. We say that $Q(s,a)$ has converged to $Q_*(s,a)$ when the difference between $Q(s,a)$ and $Q(s',a')$ is approximately equal to the transition reward *R* for any sequence $(s,a,s',R)$.

## 2.2 Reinforcement Learning Environment Configuration

We present our deep Dyna-Q learning approach by detailing its application. We outline a training procedure in the context of simulated room evacuation by a small two-wheeled robotic agent. The scenario is modeled in Gazebo, an open-source 3D robotics simulator (Koenig and Howard, 2004). Gazebo is a user-friendly graphical interface that we utilize for its high-speed simulation of collision physics in configurable 3D worlds, as well as its realistic rendering of wide-angle camera sensors on custom-designable robots.

The simulation environment is a basic Cartesian world containing a walled-in square room, an exit, and the robotic agent. The models for these objects are assembled of primitive 3D geometry elements from the Gazebo-associated SDFormat, such as `box` and `cylinder`. Models with more detail can be simulated too, but we adopt the low-polygon modeling style for the consequent improvement in computation speed. Our model of the walled-in room is constructed from four `box` elements that act as walls. The walls are 1 meter tall, enclose a square area of 2.5 x 2.5 meters, and are uniformly colored in Gazebo's default shade of blue. The exit is a single `box` element of 1 meter height, 0.5 meters width, 0.2 meters depth, and colored with Gazebo's default green. Obstacles will be introduced to the environment in a later section and will be colored with Gazebo's default red.

The agent model fits within a 0.15 x 0.15-meter footprint, and can capture images from a front-facing camera sensor configured with a wide-angle stereographic lens (**Figure 1A1**). Images are output in the R8G8B8 format at 20-pixel width by 7-pixel height (**Figure 1B**). These images are the means by which our agent observes and represents the environment, and we will refer to them as the agent's state. By this definition, the agent's state represents its incomplete perception of the true state of the environment; this implies that our proposed vision-based navigation task can be classified as a partially-observable Markov decision process.

In this work, we train an algorithm to serve as a high-level controller, analyzing the quality of a fixed-length movement trajectory rather than directly controlling motor torques. While DRL has been used to learn robust visuomotor policies (Levine et al., 2016), we assume the implementation of a feasible robotic system with preprogrammed movement routines. This simplifies the learning task while minimizing the size and depth of the neural network that is needed to approximate it.

We define the agent's action space $A$ as the set of locomotion trajectories produced by 0.5-foot displacements at relative rotation angles in [-135°, -90°, -45°, 0°, 45°, 90°, 135°] (**Figure 1A2**). In effect, the action space belonging to a state that corresponds to position $(x, y)$ and orientation $\theta$ in the environment is a set of agent location updates:

$$A = \left. \begin{array}{l} \theta' \leftarrow \theta + a \\ x' \leftarrow x + 0.5\text{ft} \cos\theta' \\ y' \leftarrow y + 0.5\text{ft} \sin\theta' \end{array} \right\} \forall\, a \in [\text{-}135°, \text{-}90°, \text{-}45°, 0°, 45°, 90°, 135°]. \quad (3)$$

## 2.3  Dyna-Q Deep Reinforcement Learning Procedure

Our code implementation of the DRL training process is written in Python version 3.6 and utilizes the TensorFlow machine learning interface (TensorFlow Developers, 2022). We use TensorFlow to design and generate an appropriate neural network to learn vision-based navigation in our environment. We devise a DQN that accepts 20x7 RGB images as input, which matches the format of the agent's state. The DQN outputs 7 numeric values, where each value indicates the perceived quality of its corresponding action $a$ in action space $A$. A notably small input size paired with a simple discretized action space allows us to use a basic fully-connected DQN, without the need for convolutional layers or recurrence. The DQN has three hidden layers that contain 64, 128, and 64 neurons respectively, each fitted with TensorFlow's built-in ReLU activation function (**Figure 1**).

In our methods, we utilize a train-target update scheme in order to stabilize the training process. This means that our procedure actually trains two DQNs, a training network and a target network. The structure and initialization of the two networks are equivalent, but the weights of the target network lag those of the training network. The training network actively learns new information during each training step, and its weights are continuously adjusted to reflect the agent's observations. In contrast, the target network is what the algorithm uses to approximate expected reward in each training step, and its weights are only updated toward those of the training network at the end of each episode.

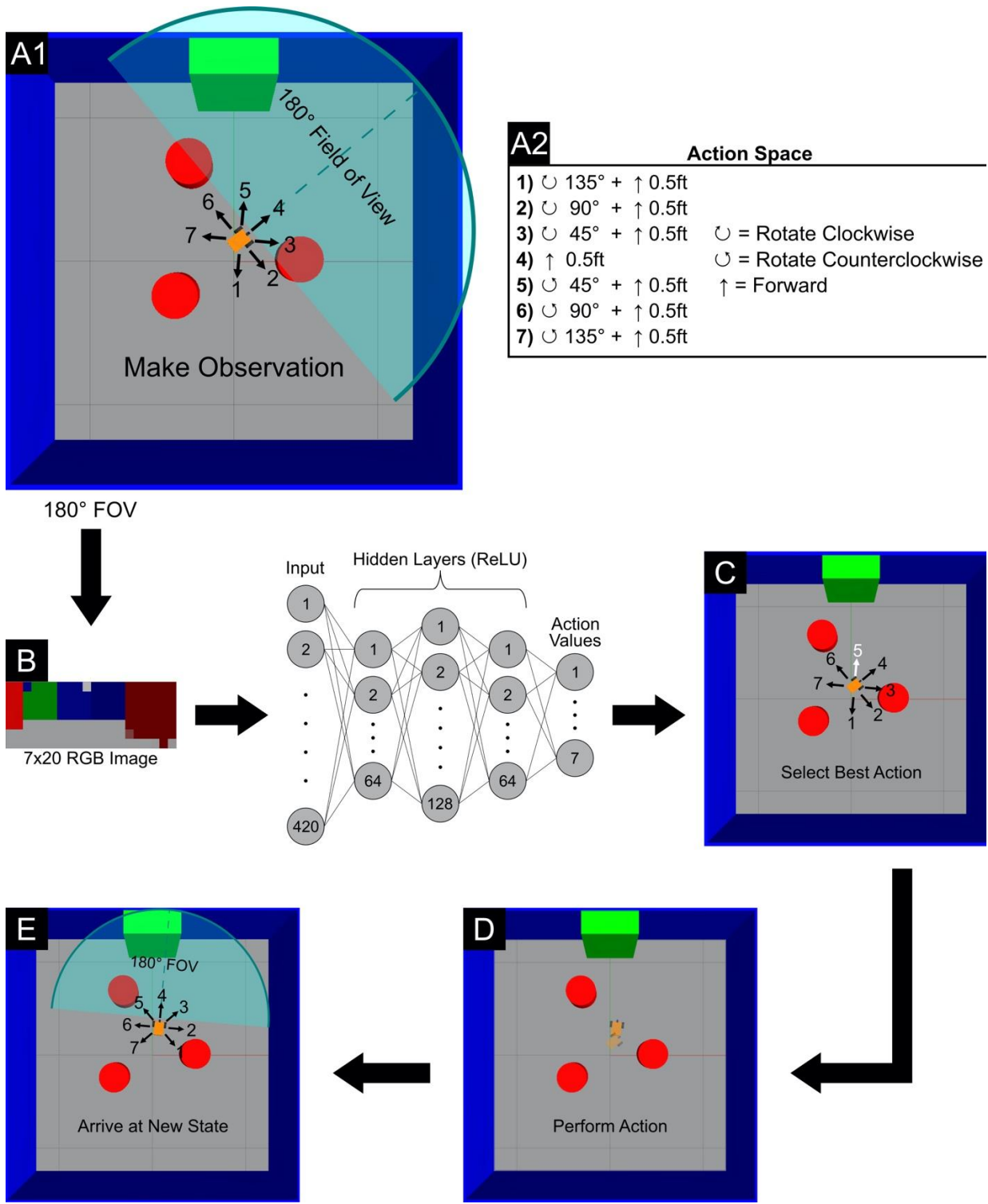

**Figure 1** Overview of the state transition process from **A**-**E**. **(A1)** The robot uses its 180° field of view camera to observe a 7x20 pixel RGB image which defines the state of the environment perceived by the agent. **(A2)** At each time step, the agent chooses one out of 7 relative rotations combined with a forward step of 0.5 meters. These compound actions comprise the action space. **(B)** The observed image is flattened from a 7x20x3 RGB matrix to a 1x420 vector and fed into the neural network. **(C)** Agent selects the action of greatest value (highlighted in white) according to the neural network. **(D)** Agent performs the selected action (rotate counterclockwise 45° and move 0.5 meters forward). **(E)** Agent arrives at the next state, makes a new observation, and then repeats the process.

As a means to train a robot for capable vision-based navigation, we implement a Dyna-Q DRL training loop (**Figure 2**). The training procedure spans 1,000 reinforcement learning episodes, each running for at most 10,000 training steps. At the start of each episode, the configuration of the environment is randomized. The exit respawns in a random position along one of the room's four walls, and the agent respawns in a random orientation at a random location within the bounds of the room, all sampled continuously. Each training episode is run either until an agent takes an action that results in successful exit and receives reward zero, or until reaching the maximal 10,000 steps, where each step has received a reward of -0.1.

We utilize the ε-greedy exploration-exploitation policy during training to induce random exploration in a variable manner. In the early stages of a training session, the agent explores its environment to generate state transition experiences that it can learn from. As the training progresses, the exploration probability (ε) decays exponentially according to:

$$\varepsilon = \varepsilon_{min} + (\varepsilon_{max} - \varepsilon_{min}) \exp\left(-\frac{4}{P}\frac{e}{e_{total}}\right), \tag{4}$$

where $\varepsilon_{min} = 0.1$ is the minimum exploration probability, $\varepsilon_{max} = 1.0$ is the maximum exploration probability, and $P = 50\%$ is the percentage of total training episodes where $\varepsilon \approx \varepsilon_{min}$. The value of $e$ increments with the current episode, and $e_{total} = 1,000$ is the total number of episodes. As the training progresses, ε-greedy will allow the agent to exploit its learned strategy and make fine-tuned adjustments to converge on the optimal policy.

As the agent takes steps through the reinforcement learning environment, it makes observations about the reward it receives consequent to its actions from specific states. Explicitly, we can say that at time $t$ the agent observes state $s_t$, chooses an action $a_t$ informed by its perception of $s_t$, then transitions to state $s_{t+1}$ and receives reward $R_{t+1}$ (**Figure 1C-E**). A new observation about the environment is made every time the agent takes a step from time $t$ to $t+1$, encapsulated by the experience tuple ($s_t$, $a_t$, $s_{t+1}$, $R_{t+1}$). Each observation ($s_t$, $a_t$, $s_{t+1}$, $R_{t+1}$) is stored in a memory buffer and will be reconsidered multiple times before it is discarded. The memory buffer is implemented as a First-In-First-Out (FIFO) queue with a maximum length of 10,000 experiences.

The key insight of the Dyna-Q algorithm is the use of stored memories for experience replay, a form of model-based planning that helps to reduce variance and prevent overfitting. Accordingly, experience replay is performed every time the agent takes a step in our training process. The experience replay method utilizes Q-learning to update the weights of the training DQN, as per equation (2), to reflect the information that is learned from a randomly sampled batch of 50 experiences from the memory buffer. We configure the Q-learning process for an experience tuple ($s_t$, $a_t$, $s_{t+1}$, $R_{t+1}$) using a loss function that corresponds to the mean squared temporal difference error:

$$L(w_{train}) = \left(R + \gamma \max_{a'} Q(s', a'; w_{target}) - Q(s, a; w_{train})\right)^2, \tag{5}$$

where $w_{target}$ and $w_{train}$ indicate that Q-value approximation uses weights of the target DQN and training DQN, respectively. This loss function is used to refine the training DQN with every agent step, calling upon TensorFlow's built-in Adam Optimizer (Kingma and Ba, 2017) to tune the weights $w_{train}$. In our methods, the learning rate $\alpha$ is set to $1\times10^{-4}$ and the discount factor γ equals 0.999.

At the end of each training episode, the weights of the target network are updated towards those of the training network by a factor of $\tau$:

$$w_{target} \leftarrow w_{target} + \tau(w_{train} - w_{target}). \quad (6)$$

We choose a target update factor of $\tau = 0.1$, inspired by a successful application of the same factor to smooth the learning curve without sacrificing learning speed (Zhang et al., 2021).

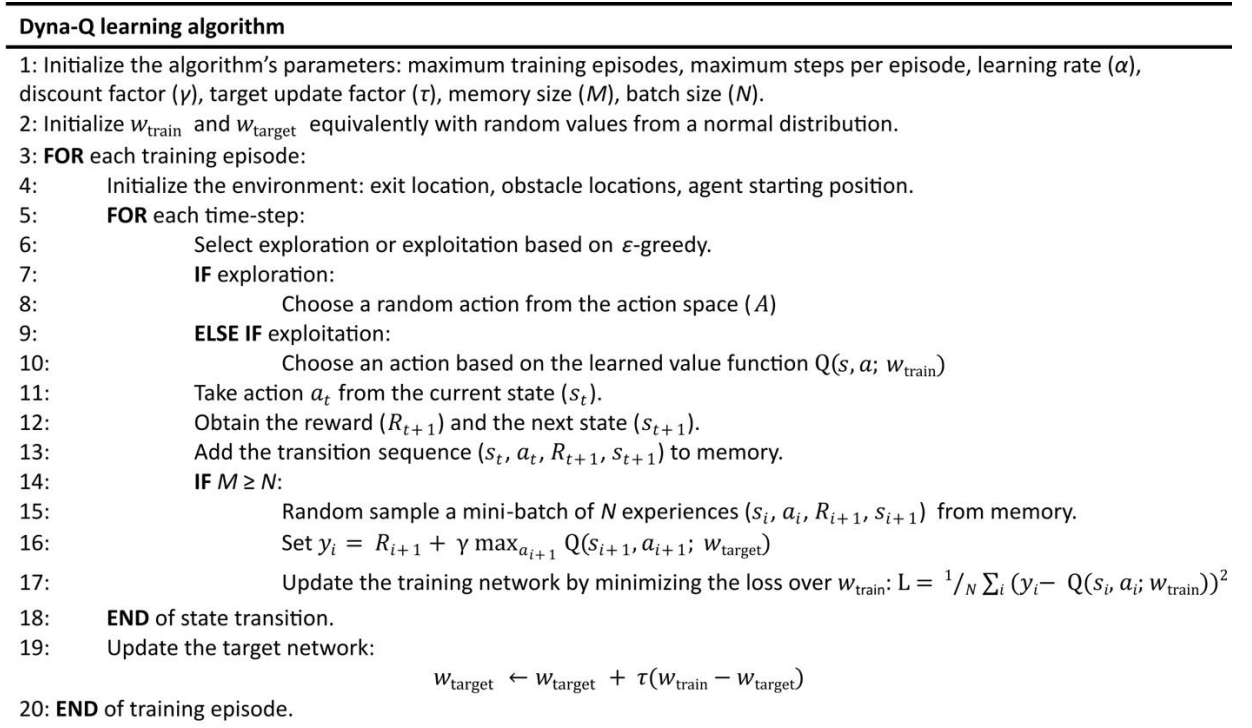

**Figure 2** Pseudocode outlining implementation of the deep Dyna-Q algorithm.

During its traversal through the reinforcement learning environment, the agent should refrain from touching obstacles or the walls of the room. We deter these collisions by invalidating them during training. If the agent chooses to take an action that would result in a collision, it instead stays in place for the step and perceives an identical image state while still receiving a -0.1 reward.

In this work, we use a custom-developed Gazebo WorldPlugin to detect when the agent collides with an obstacle or with the wall. To optimize runtime, our plugin bypasses Gazebo's message-passing communication system to directly scan the underlying physics engine for contact points on the agent collision body. The plugin also checks whether any detected contact points belong to the exit collision body, which determines evacuation status for the episode.

The proposed DRL training process is discretized to work in steps, and it requires the agent to capture a single image from the environment exactly one time per training step. For an extreme improvement in simulation speed, our implementation makes the same discretization in the computation process. The plugin uses manual simulation stepping to achieve a minimal number of wasted executions. We skip the simulation of robotic locomotion, as this is not efficient during the learning process. Instead, on each training step we use our plugin to update model positions and then simulate a single iteration of the collision physics and graphics rendering in Gazebo.

## 3   Results

We demonstrate the effectiveness of the proposed DRL methods by applying them in a series of experiments in the context of an evacuation scenario. As described in the previous section, our methods are based on a virtualized training process built in Gazebo and Python. In this section, we demonstrate the application of this process with a simulated two-wheeled robotic agent in a variety of evacuation experiments. We show that the algorithm can consistently converge to an optimal Q-value approximator in a relatively small amount of computation time, even with minimal input data and a simple neural network design. Randomization of the environment between episodes encourages the development of an optimal strategy that can demonstrate generalized situational understanding of its surroundings in unpredictable configurations.

During training, we use an ε-greedy exploration strategy to induce further randomness in the agent's trajectories. At least 10% of the actions taken in any training episode are selected randomly. While testing a trained algorithm, random exploration is disabled and the agent follows its learned policy exactly. The following sections present such test results in a series of room evacuation experiments. Successive experiments build on one another in difficulty as various obstacles of differing complexity are introduced.

### 3.1   Room Evacuation

In this experiment, we use Gazebo to simulate a basic room evacuation scenario. This simulated world serves as the reinforcement learning environment for our experiments (depicted from a top-down view in **Figure 4**). Recall that the purpose of our proposed methods is to teach an agent to navigate to a target (the exit) in the fewest possible number of fixed-sized steps. We apply these methods in the simulated reinforcement learning environment and show that our agent learns to skillfully navigate a series of 0.5-foot steps to the exit.

At a high level, our implementation of the environment in this experiment provides two functions that are utilized in the training process: reset and step. Calling the reset function in this environment randomizes its configuration, preparing the simulation for a reinforcement learning training episode. The exit respawns in a random position along one of the room's four walls, and the agent respawns at a random location within the bounds of the room. A single frame is then captured from the agent's onboard camera and rendered as a 20x7 pixel image to be returned to the caller. This image is the first state in the episode, ideally unique due to the nature of the random environment configuration. The environment also implements the step function, which requires passing the index of an action from the action space as a parameter. The step function moves the agent one step-length from its current position at a relative angle that is determined by the selected action. Similarly to reset, the step function returns the latest image from the agent's camera to update the state of the training episode.

Simulating our training procedure in this environment took 19 minutes and 2.2 seconds of computation time on a 2017 MacBook Pro with a 2.3 GHz Dual-Core Intel Core i5 processor and 8 GB RAM. The agent executed a total of 51,316 steps over 1,000 training episodes. We observe that the majority of steps are taken in earlier training episodes, and that the total number of steps taken per episode decreases exponentially in response to agent learning. Correlating to our negative reward formulation, cumulative episodic reward increases toward optimality over the course of the training session (**Figure 3**).

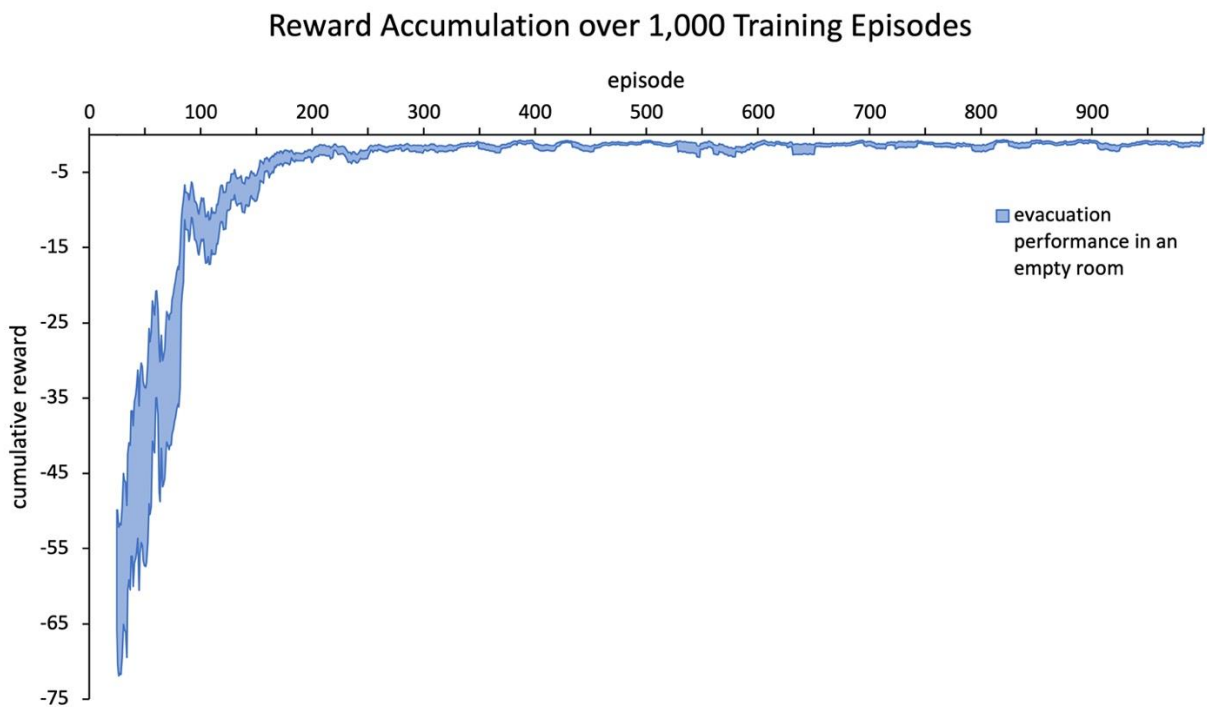

**Figure 3** Performance evolution during a training session for empty room evacuation. The cumulative episodic reward converges to a maximum value as training progresses.

In testing the trained algorithm, we find that following the learned policy leads to efficient room evacuation in arbitrary exit configurations, and we observe competitive trajectories that consistently end with successful escape. We illustrate these results in **Figure 4**, where three sampled escape paths are mapped through the reinforcement learning environment. Each arrow in these paths corresponds to a single step in their traversal. For illustration purposes the exit is kept stationary in a central location during all three episodes, but it's worth noting that evacuation performance would not be affected if the exit were positioned elsewhere. The figure exhibits the agent's perceived state at three random starting locations. These states are 20x7 pixel images that are input directly into the trained algorithm to estimate the seven consequent Q-values. We expect trained Q-values to be negative, so for clearer illustration we consider their softmax. The softmax function normalizes the Q-values so that they can be interpreted as a probability distribution over the action space (Goodfellow et al., 2016, 180–184). Action preference distributions corresponding to each depicted state are plotted as probabilities on polar axes in the figure, while actual Q-values are encoded by color for reference.

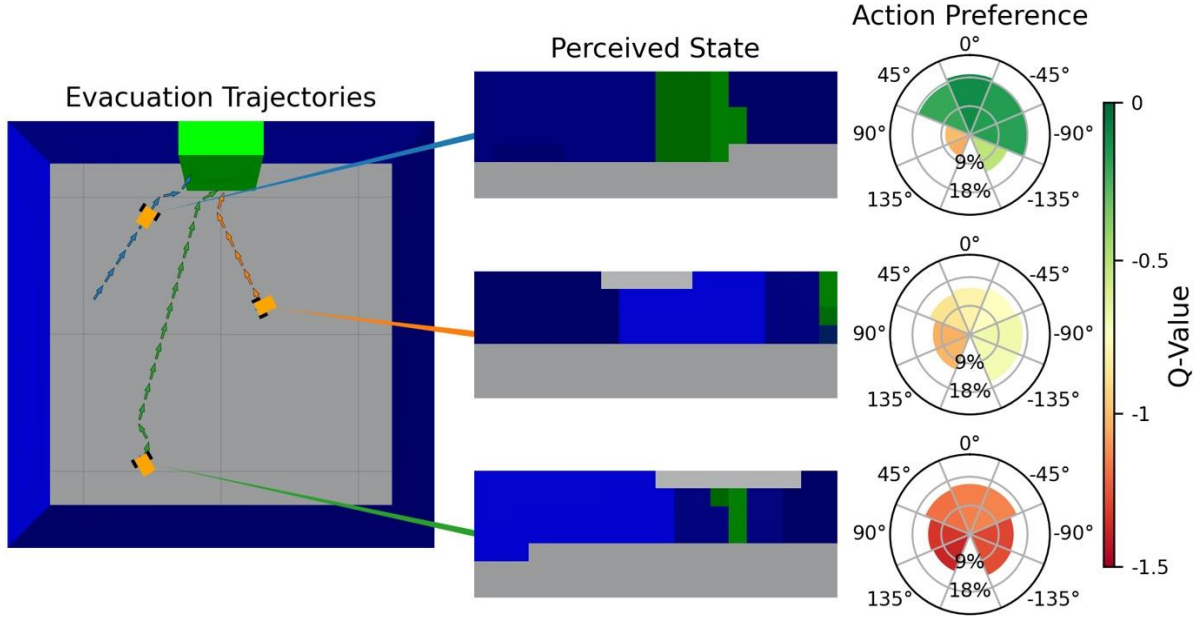

**Figure 4** Evacuation trajectories in an empty room are shown. The agent is initialized in random locations and navigates to the exit. The environmental state is perceived as 20x7 pixel RGB images from an onboard camera; three sampled state perceptions are depicted at center. The algorithm learns to assess action values, developing a preference toward optimal actions. Learned preference is illustrated at right, where action partiality at the sampled states is plotted as a percentage versus relative locomotion angles in the action space. Corresponding Q-values are encoded by color in the plots.

## 3.2 Room Evacuation with Static Obstacle Avoidance

### 3.2.1 Avoiding Cylindrical Obstacles in Arbitrary Configurations

Building upon the previous experiment, we utilize our DRL procedure to teach camera-based obstacle avoidance in a simulated evacuation situation where static obstacles are introduced to the environment. An obstacle is modeled as a vertically-oriented `cylinder` geometry element with 0.5-foot radius, 0.3 meter height, and Gazebo's default red coloring. Multiple obstacles of this type can be spawned into the reinforcement learning environment at arbitrary locations (**Figures 5, 6**). Whenever the environment is reset, the obstacles are respawned at random positions within a centrally located 1.4 x 1.4 meter square area. The reset function ensures that obstacle positions are separated from one another by at least 0.7 meters and from the agent by at least 0.5 meters during its initialization of each training episode.

We've simulated our methods in a progression of obstacle avoidance experiments that study evacuation in the presence of up to three obstacles. We find that our procedure produces proficient strategies that demonstrate confident navigation through an increasingly cluttered space of randomly scattered obstacles. Training results reveal a reasonable increase in required step executions as more obstacles are introduced to the environment: our 1,000-episode training process took 41,687 steps with one obstacle and 59,108 steps with three obstacles. Computation time remained moderate, culminating at 36 minutes and 53.4 seconds in the case of three static obstacles.

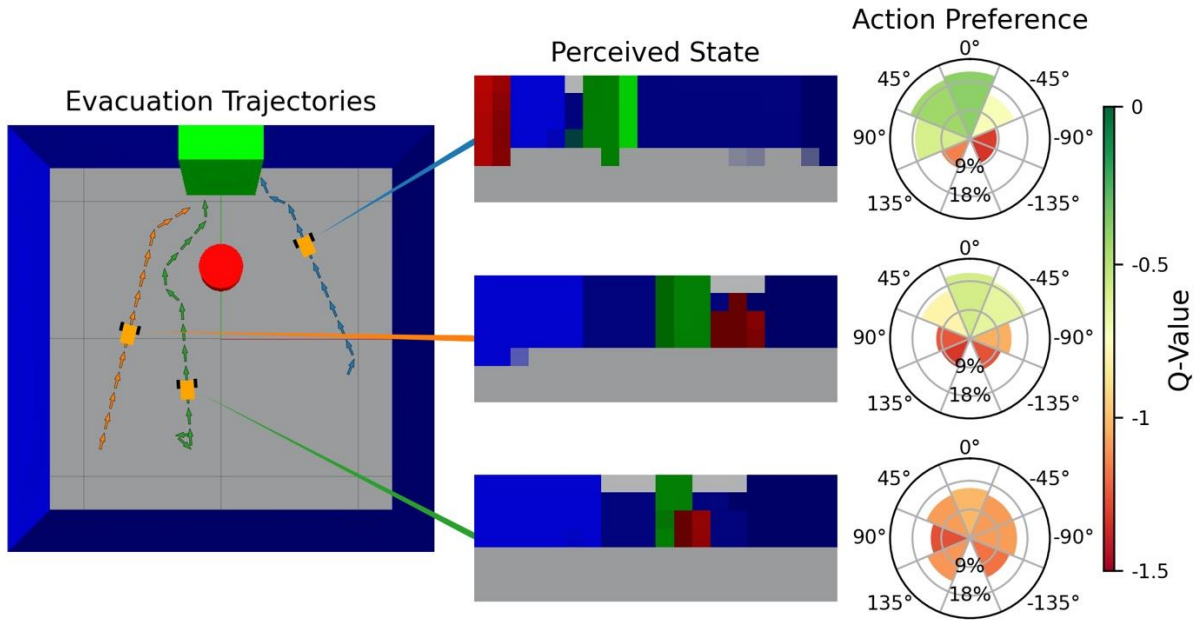

**Figure 5** Illustration of evacuation trajectories in the presence of a static obstacle. The agent is initialized in random locations and navigates to the exit. The environmental state is perceived as 20x7 pixel RGB images from an onboard camera; three sampled state perceptions are depicted at center. The algorithm learns to assess action values, developing a preference toward optimal actions. Learned preference is illustrated at right, where action partiality at the sampled states is plotted as a percentage versus relative locomotion angles in the action space. Corresponding Q-values are encoded by color in the plots.

Examination of the resulting policies confirms appropriate adaptation to obstacles, and we observe that the trained networks exhibit stronger preference toward individual actions along with increased variance in estimated Q-values. We explore this result in **Figure 5**, where sampled evacuation paths are pictured alongside featured state perceptions and corresponding action preferences. For illustrational simplicity this figure depicts an unchanging evacuation scenario with a single obstacle. We note unfaltering performance in the presence of up to three obstacles and with arbitrary environmental configuration, where testing demonstrates that the agent can traverse around and between multiple obstacles (**Figure 6**).

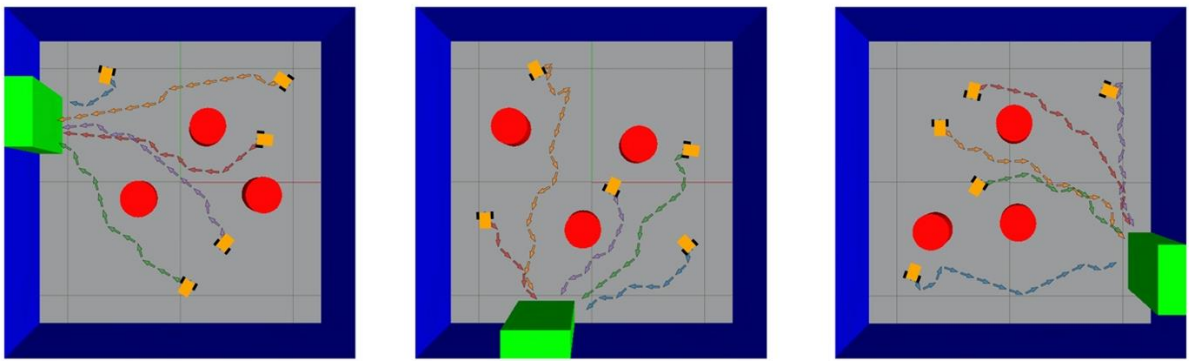

**Figure 6** Evacuation trajectories in randomized environments are shown. A vision-based strategy allows the agent to navigate and avoid obstacles in unpredictable configurations.

### 3.2.2 Avoiding a Concave Obstacle

Previous literature has revealed that many traditional path planning algorithms can be thwarted by concave-shaped obstacles. DRL methods appear advantageous in this regard, evidenced by a proven capability to avoid trapping by such obstructions (Zhang et al., 2021). We support this claim by demonstrating capable evacuation in the presence of a concave obstacle.

Returning to our Gazebo simulation, we install a large concave obstacle in the center of the reinforcement learning environment. The obstacle is modeled using a `polyline` geometry element that generates a curved wall; it is 0.1 meters thick and 0.25 meters tall, and it traces a smooth semicircular arc that opens to a 1.2 meter diameter. The position of the concave obstacle does not change during this experiment, whereas the positions of the exit and the agent are randomized on reset as per previous procedure. We find that the agent learns a versatile strategy that showcases skillful evacuation versus the concave obstacle, even with arbitrary exit configurations. **Figure 7** highlights sampled escape paths in a notably difficult exit configuration, demonstrating successful evacuation in spite of the presented concavity. Featured state perceptions illustrate the inherent complexity of this path-planning problem, while corresponding action preferences show convergence to a capable navigation policy. Simulating 1,000 training episodes in this environment took 58 minutes and 38.8 seconds, completing the learning process in 113,839 steps. Randomization of the environment between training episodes encouraged thorough exploration of its inherent rules, providing sufficient experience for the algorithm to develop a fine-tuned understanding that avoids trapping by the concave obstacle.

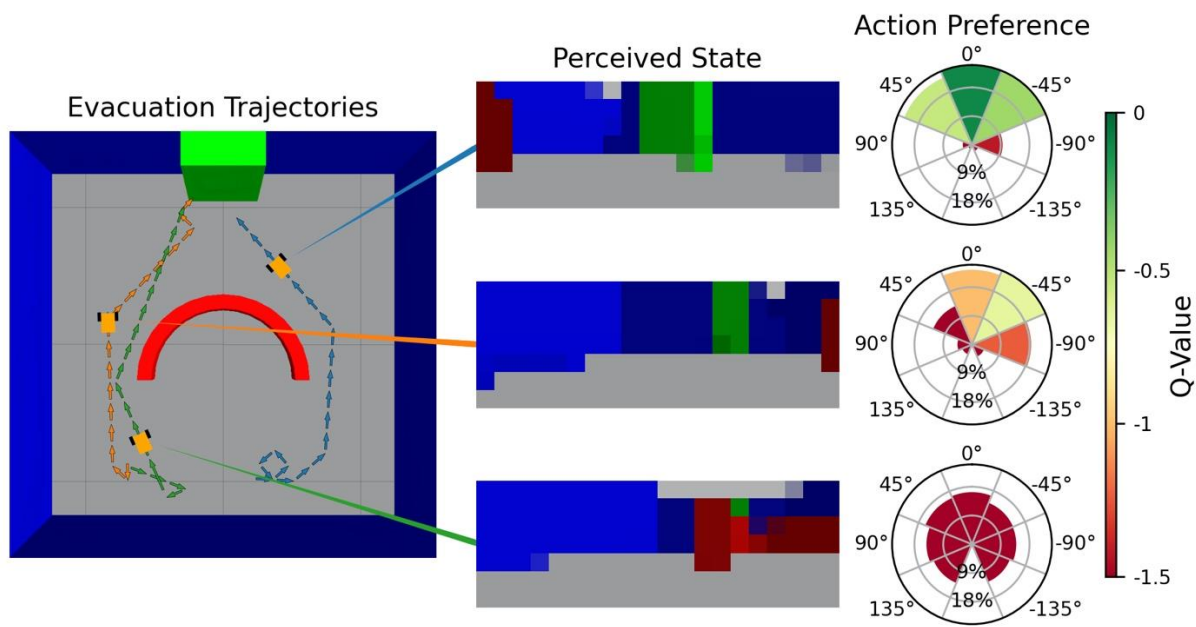

**Figure 7** Illustration of evacuation trajectories in a room with a concave obstacle. The agent is initialized in random locations and must navigate to the exit. The environmental state is perceived as 20x7 pixel RGB images from an onboard camera; three sampled state perceptions are depicted at center. The algorithm learns to assess action values, developing a preference toward optimal actions. Learned preference is illustrated at right, where action partiality at the sampled states is plotted as a percentage versus relative locomotion angles in the action space. Corresponding Q-values are encoded by color in the plots.

## 3.3 Room Evacuation with Dynamic Obstacle Avoidance

One of the most attractive justifications for our study of camera-vision is in empowering the perception of high-dimensional information that may be encoded by visual elements and color patterns. In this experiment, we investigate our algorithm's use of this perception to adapt to dynamic environments where obstacles are in continuous motion. We demonstrate that the application of our DRL procedure generates capable navigation policies in contempt of multiple dynamic obstacles, which is notably difficult for many modern algorithms that use kinetics-based perception.

The dynamic obstacle presented here is based on the cylindrical obstacle that was introduced in **3.2.1**. The 3D model remains exactly the same, but in this experiment dynamic obstacles are configured to move in randomized trajectories around the room while the agent evacuates. Multiple obstacles of this type can be spawned into the reinforcement learning environment at arbitrary locations, but successive locations are self-imposed and unpredictable. As these trajectories introduce inherent randomness in every training step, dynamic obstacle positions are not randomized on reset as static cylindrical obstacles were.

Dynamic obstacles move similarly to the agent; every time the agent takes a step in this environment dynamic obstacles will also take one step. The relative movement direction for each of a dynamic obstacle's steps is selected according to a normalized random distribution, which produces more natural-looking movement behavior. Dynamic obstacles take 0.025 meter steps, and are bounded to a 1.5 x 1.5 meter square area in the center of the environment. If multiple dynamic obstacles are present in an environment, they will stay at least 0.5 meters apart from one another such that the agent can always pass between them.

We've simulated our methods in a progression of dynamic obstacle avoidance experiments, studying evacuation with up to three moving obstacles. Our DRL procedure demonstrates repeated success in solving the room evacuation problem with dynamic obstacles. We find that agents trained in this environment can exhibit intelligent decision making that conveys time-dependent consideration of the navigation task.

## 4 Discussion

In this paper, we develop a novel deep Dyna-Q learning approach for versatile target-driven navigation and obstacle avoidance based on low resolution monocular camera images. Our reward formulation trains a DQN to assign greater values on actions that direct an agent to the exit in the shortest possible time. In this way, agents can efficiently evacuate a variety of static and dynamic environments containing both convex and concave obstacles by selecting actions of greatest cumulative value according to the trained DQN. Our results show that the agent can explore the evacuation space until it identifies the location of the exit with its camera. At the same time, it can identify areas populated with red obstacles as undesirable, as they are prone to collisions which can hinder the evacuation process. Thus, the agent employs the optimal strategy of surveying the environment until it identifies the exit, and then moving towards the exit while making small adjustments to navigate around obstacles. If the exit is obscured from the agent's view by obstacles, the robot will continue to search for the exit while maintaining a collision-free configuration. Initializing the exit and obstacles at different random locations at the start of each episode helps to generalize experience in order to generate a robust evacuation policy.

Environments with elements in motion, other than the agent, are not often considered in modern DRL navigation algorithms. Therefore, the ability of our algorithm to avoid multiple dynamic obstacles is an important contribution to the field of intelligent robotic navigation. We demonstrate that a robotic agent trained with our Dyna-Q-based approach can evacuate an environment with multiple dynamic obstacles just as efficiently as it would a static environment. We also demonstrate the agent's ability to avoid and escape from a large concave obstacle, which is an important skill that often eludes traditional path planners.

Our algorithm and methods can be easily augmented using a variety of machine learning techniques. Convolutional layers could be incorporated to give our system the ability to deal with more visually realistic environments as in (Zhu et al., 2017) and (Kulhánek et al., 2019). Additionally, we believe that research into preservation of task efficacy across different trainings using successor-feature-based learning, as in (Zhang et al., 2017), will be an important direction for this and other DRL algorithms. Overall, our generalized learning approach based on Dyna-Q is a powerful tool for visual-based start-to-goal navigation in partially observable dynamic environments.

## 5     Author Contributions

PB, PC, and GL significantly contributed to conceptualization and development of the presented approach and methods. PB and PC implemented the deep reinforcement learning (DRL) code and made figures. PB wrote the code that introduces Gazebo functionalities to the DRL code. PB made the videos. PB, PC, and GL contributed to writing and reviewing this manuscript.

## 6     Acknowledgments

The authors would like to thank Yihao Zhang and Zhaojie Chai for meaningful discussions in relation to DRL implementation.